\documentclass[11pt]{article}

\usepackage[final]{acl}   

\usepackage{times}
\usepackage{latexsym}
\usepackage[T1]{fontenc}
\usepackage[utf8]{inputenc}
\usepackage{microtype}

\usepackage{inconsolata}
\usepackage{url}
\usepackage{hyperref}

\usepackage{amsmath}
\usepackage{amsthm}
\usepackage{amsfonts}
\usepackage{amssymb}
\usepackage{nicefrac}
\usepackage{xcolor}

\usepackage{graphicx}
\usepackage{booktabs}
\usepackage{multirow}
\usepackage{caption}

\usepackage{algorithm}
\usepackage{algorithmic}

\usepackage{enumitem}

\let\bbb\textbf
\newcommand{\system}{MCPAgentBench}

\newcommand{\premcpservernum}{9714}
\newcommand{\premcptoolnum}{over 20,000}

\title{\system: A Real-world Task Benchmark for Evaluating LLM Agent MCP Tool Use}

\author{
\textbf{Zixiang Liu}\textsuperscript{1}\thanks{Zixiang Liu, Wenrui Liu, and Elsie Dai contributed equally to this work.},
\textbf{Wenrui Liu}\textsuperscript{1,*},
\textbf{Elsie Dai}\textsuperscript{1,*},
\textbf{Wenhan Yu}\textsuperscript{1}\\
\textbf{Lei Yu}\textsuperscript{1},
\textbf{Tong Yang}\textsuperscript{1}\thanks{Corresponding author: \texttt{yangtong@pku.edu.cn}},
\textbf{Jinjun Han}\textsuperscript{2},
\textbf{Hong Gao}\textsuperscript{2}
\\
\textsuperscript{1}Peking University
\qquad
\textsuperscript{2}ZTE
}

\begin{document}

\maketitle

\begin{abstract}


Large Language Models (LLMs) are increasingly serving as autonomous agents, and their utilization of external tools via the Model Context Protocol (MCP) is considered a future trend. Current MCP evaluation sets suffer from issues such as reliance on external MCP services and a lack of difficulty awareness. To address these limitations, we propose \system{}, a benchmark based on real-world MCP definitions designed to evaluate the tool-use capabilities of agents. We construct a dataset containing authentic tasks and simulated MCP tools. The evaluation employs a dynamic sandbox environment that presents agents with candidate tool lists containing distractors, thereby testing their tool selection and discrimination abilities. Furthermore, we introduce comprehensive metrics to measure both task completion rates and execution efficiency. Experiments conducted on state-of-the-art LLMs reveal significant performance differences in handling complex, multi-step tool invocations. All code is open-source at \cite{MCPAgentBench}.
\end{abstract}

\section{Introduction}

Large Language Models (LLMs) \cite{vaswani2017attention,brown2020language,deepseekai2025deepseekr1,yang2025qwen3} achieve breakthrough progress in natural language processing and complex reasoning tasks. To further advance capabilities, the research community shifts focus toward Agent models \cite{park2023generative,wang2023survey,xi2023rise,hong2023metagpt}. The Model Context Protocol (MCP) \cite{modelcontextprotocol_intro_2025,hou2025mcp,ehtesham2025agentprotocols,lumer2025scalemcp} currently represents a crucial exploration aimed at unifying the interaction modality between Agents and external tools, defining a standard format for tool invocation. Agent utilization of MCP tools becomes a key approach for solving complex, real-world tasks. Consequently, establishing a comprehensive evaluation benchmark that assesses the planning and execution capabilities of Agents in invoking MCP tools is essential.

However, existing MCP capability assessment benchmarks \cite{luo2025mcpuniverse,gao2025mcpradar,fan2025mcptoolbench} suffer from several significant limitations. Firstly, a stability and dependency issue exists, as current benchmarks often rely on real, remote MCP servers, where service stability and availability heavily impact the reproducibility of testing results. Secondly, these benchmarks exhibit insufficient difficulty awareness, performing only simple task categorization and lacking granular observation of the invocation complexity level. Most importantly, models should be able to efficiently complete tasks, while existing frameworks lack metrics for task execution efficiency. 
To address these challenges, a new evaluation benchmark must incorporate: local MCP server deployment to ensure stability; comprehensive coverage of complex invocation scenarios, including single-step, serial, and parallel calls; and dedicated metrics for task execution efficiency.

To achieve these goals, this paper proposes \system{}, an evaluation benchmark specifically designed to assess the efficiency of Agent MCP tool invocation locally.
This evaluation work makes the following contributions:
\begin{itemize}
    \item Local Deployment of MCP Servers Enabling Easily Reproducible Evaluation: We collect \premcptoolnum{} MCP Tools from sources including MCP Marketplace \cite{mcpmarket_cn_2025} and HuggingFace \cite{alihmaou_2025_agents_mcp_hackathon}, and then perform reconstruction of selected tools into 141 locally maintained MCP servers to provide controlled execution settings as well as reproducible evaluation while preserving authentic tool definitions and protocol-compliant behavior.
    \item A Structured Task Suite Grouped by Invocation Complexity: 841 raw tasks in total are collected from datasets including Infinity-Instruct~\cite{infinityinstruct_dataset_2025} and Schema-Guided Dialogue Dataset~\cite{rastogi2020scalablemultidomainconversationalagents}. They are organized according to invocation complexity and task characteristics, allowing systematic analysis across different categories of tool-use difficulty. 178 high-quality tasks are finally constructed through manual labeling and matching between tasks and tools.
    \item Comprehensive Metrics for Accuracy and Efficiency Evaluation: The Task Finish Score (TFS), Task Efficiency Finish Score (TEFS),Time Efficiency, and Token Efficiency metrics adapt widely used agents’ evaluation paradigms to MCP tool invocation, enabling quantitative analysis of Agent’s planning correctness, execution timing, and resource consumption under a unified protocol. This compatibility facilitates comparison with prior agent evaluation while providing additional insight into the efficiency of MCP-based tool use.
\end{itemize}


\section{Related Work}
\label{sec:related-work}
 \begin{figure*}[h]
    \setlength{\abovecaptionskip}{0.1cm}
    \setlength{\belowcaptionskip}{-0.35cm}
    \centering  
\includegraphics[width=1\linewidth]{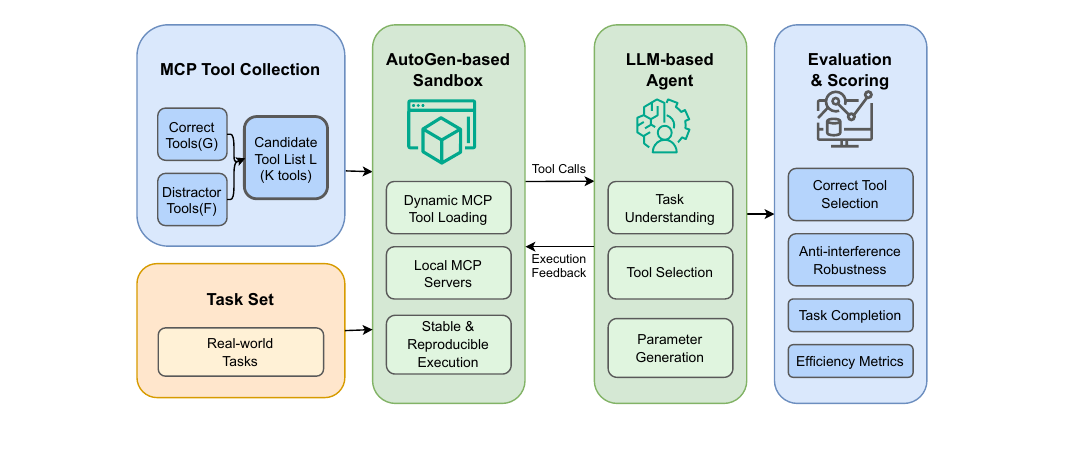}
    \caption{\system{} Overview.}
    \label{fig:overview}
\end{figure*}

Recent advances in agentic large language models (LLMs) have shifted evaluation from pure text generation toward tool-use reasoning, with a growing emphasis on when, how, and why agents invoke tools. The Model Context Protocol (MCP) has emerged as a unified substrate for such evaluation by enforcing schema-consistent communication and execution-grounded validation, enabling reproducible and protocol-aligned assessment of tool-augmented agents.

Before MCP, tool-use benchmarks were developed largely within heterogeneous API ecosystems. API-Bank~\cite{Li2023APIBank} and ToolBench~\cite{Qin2024ToolLLM} evaluated Plan--Retrieve-Call behaviors but lacked unified protocol abstraction and execution-level guaranties. Earlier paradigms such as ReAct~\cite{Yao2023ReAct}, Auto-GPT~\cite{Richards2023AutoGPT}, and GAIA~\cite{mialon2023gaia} explored the interaction between reasoning and acting, though often in synthetic or text-only environments. More recent MCP-based benchmarks represent a clear shift from simulated function calling to execution-verified, protocol-consistent evaluation. MCP-Universe~\cite{luo2025mcpuniverse} and MCP-Bench~\cite{wang2025mcpbench} evaluate agents against live MCP servers, MCP-RADAR~\cite{gao2025mcpradar} introduces multi-dimensional evaluation metrics, MCPWorld~\cite{yan2025mcpworld} supports hybrid API--GUI tasks, LiveMCPBench~\cite{mo2025livemcpbench} evaluates agents’ retrieval and multi-tool composition abilities in large-scale MCP ecosystems, and MCPToolBench++~\cite{fan2025mcptoolbench} further scales evaluation to thousands of MCP servers with a fine-grained error taxonomy.

While these benchmarks substantially advance the realism and coverage of MCP-based evaluation, they primarily focus on task correctness and protocol compliance. In contrast, MCPAgentBench is proposed as a complementary data set that targets a more fine-grained and decision-centric aspect of tool use: the efficiency with which agents select and invoke MCP tools to complete tasks. In terms of concreteness, MCPAgentBench differs from the existing MCP benchmarks in several key aspects. It employs an Autogen-based sandbox with locally maintained MCP servers to ensure stable and reproducible execution. Tasks are categorized according to the complexity of MCP tool invocation and task attributes, enabling structured analysis across difficulty levels. Moreover, MCPAgentBench uses authentic definitions and parameters of the MCP tool, building simulated MCP servers that strictly follow the standard MCP protocol, while introducing realistic distractors that test the’ robustness of agents in tool selection. Together, these design choices position MCPAgentBench as a complementary benchmark that emphasizes task completion efficiency, task authenticity, and robustness to interference, allowing fine-grained evaluation of agent tool-invocation capabilities beyond correctness alone.




\section{\system{} Framework}
This section introduces the architecture of \system{}, the benchmark construction process, the evaluation process, task classification, and evaluation metrics.

\subsection{Overview}
 \begin{figure*}[h]
    \setlength{\abovecaptionskip}{0.1cm}
    \setlength{\belowcaptionskip}{-0.35cm}
    \centering  
\includegraphics[width=1\linewidth]{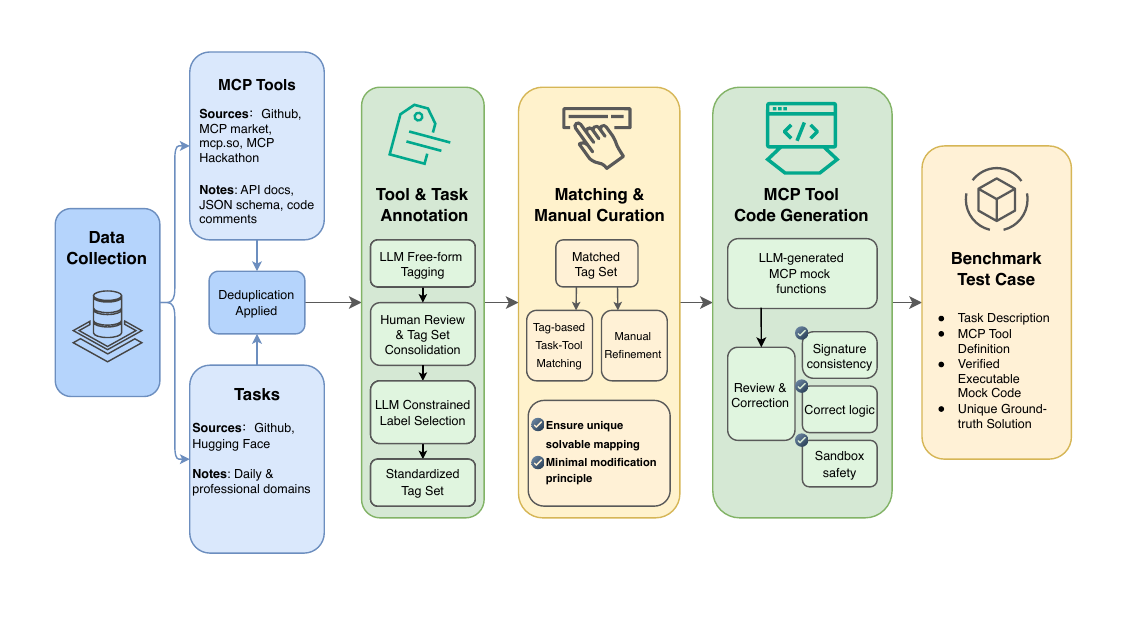}
    \caption{The Data Preprocess of \system{}.}
    \label{fig:datapreprocess}
\end{figure*}

\system{} employs a sandbox environment built upon the Autogen framework~\cite{wu2023autogen}. For each task, the system dynamically loads a corresponding MCP tool list via Autogen's tool interface to facilitate automated benchmark testing.

The overall architecture of the \system{} framework, illustrated in Figure \ref{fig:overview}, comprises three key components:

\begin{itemize}[leftmargin=*, labelsep=0.5em]
\item MCP Tool Collection: A repository of authentic MCP tools, collected and curated from GitHub and various MCP collection websites. 
\system{} extracts structured information for each tool, particularly its functional description and parameters.

\item Task Set: A diverse collection of tasks spanning daily life and professional domains. These tasks, originating from real-world datasets, have undergone meticulous manual review and curation to ensure the uniqueness of each task's solution.

\item Automated Evaluation Sandbox: A sandbox environment implemented using the Autogen framework, designed for automated task execution and evaluation.

\end{itemize}
The evaluation process leverages this sandbox environment. For each task $T$, \system{} retrieves $n$ corresponding correct tools ($G$) from the main tool library and samples $K-n$ "distractor tools" ($F$) that are functionally unrelated or easily confused. Together, these form a dynamic candidate list $L$ containing $K$ tools (e.g., $K=20, 30$), which \system{} provides to the agent under test at runtime.

The agent (driven by the LLM under test) must interpret task $T$, select the correct tool(s) from the distractor-filled list $L$, and generate compliant call parameters. The Autogen framework manages the agent-tool interaction and records every tool call. Finally, \system{} compares these captured calls against the pre-defined, unique solution and automatically computes a score based on the evaluation metrics.
During comparison, \system{} by default compares both the name of the called MCP tool and the incoming parameters with the solution. For tools where parameters are not unique, \system{} only compares whether the names are consistent.

This design not only tests the model's fundamental tool-calling capabilities but also specifically assesses its tool discrimination and anti-interference abilities in a "needle in a haystack" scenario. Users can initiate the fully automated evaluation simply by providing an API key and model name in the configuration file.

The Autogen-based sandbox provides a unified environment for evaluation. The benchmarking process utilizes 178 human-curated test cases, which load sequentially into the sandbox for testing. Furthermore, \system{} supports user-defined test cases, allowing for the expansion of the evaluation scope by following the specified instance format.

\subsection{Data Preprocessing}

The quality of the benchmark hinges on the authenticity and rigor of its data. To construct high-quality test cases, \system{} employs a four-step data processing workflow designed to ensure task authenticity, tool representativeness, and solution uniqueness.

\bbb{Step 1}: Raw Data Collection. 
As shown in Figure \ref{fig:datapreprocess}, the process begins by collecting raw data from two primary public channels. 
For MCP Tools, we gather authentic Model Context Protocol servers and tool definitions from various websites, including \cite{punkpeye2025awesome},  \cite{mcpmarket_cn_2025}, \cite{mcp_so}, and \cite{alihmaou_2025_agents_mcp_hackathon}. 
These definitions typically exist in the form of API documentation, JSON Schemas, or code comments. 
After deduplication, we obtain definitions for \premcpservernum{}  MCP servers and \premcptoolnum{} MCP tools. For Tasks, we collect real-world user queries and task descriptions from the Hugging Face Datasets platform and other academic datasets like \cite{infinityinstruct_dataset_2025} and  Schema-Guided Dialogue Dataset \cite{rastogi2020scalablemultidomainconversationalagents}.
These tasks cover both daily and professional domains.

\bbb{Step 2}: Tool and Task Annotation. The collected raw tools and tasks are functionally disparate. To establish connections between them and ensure label standardization, \system{} utilizes a three-stage, LLM-based annotation process. First, this stage leverages the open-ended generation capability of an LLM to perform an initial, unconstrained, free-format annotation of all MCP tools and tasks, allowing the model to generate multiple descriptive labels for each item. 
Next, this is followed by manual integration and filtering of all generated tags. This process aims to merge synonyms, remove ambiguous labels, and establish a unified, standardized "Tag Set". Finally, an LLM is utilized again, but constrained to select labels only from this established Tag Set. The model selects the most appropriate label(s) for each MCP tool and task, ensuring consistent classification.
 \begin{figure*}[h]
    \setlength{\abovecaptionskip}{0.1cm}
    \setlength{\belowcaptionskip}{-0.35cm}
    \centering  
\includegraphics[width=1\linewidth]{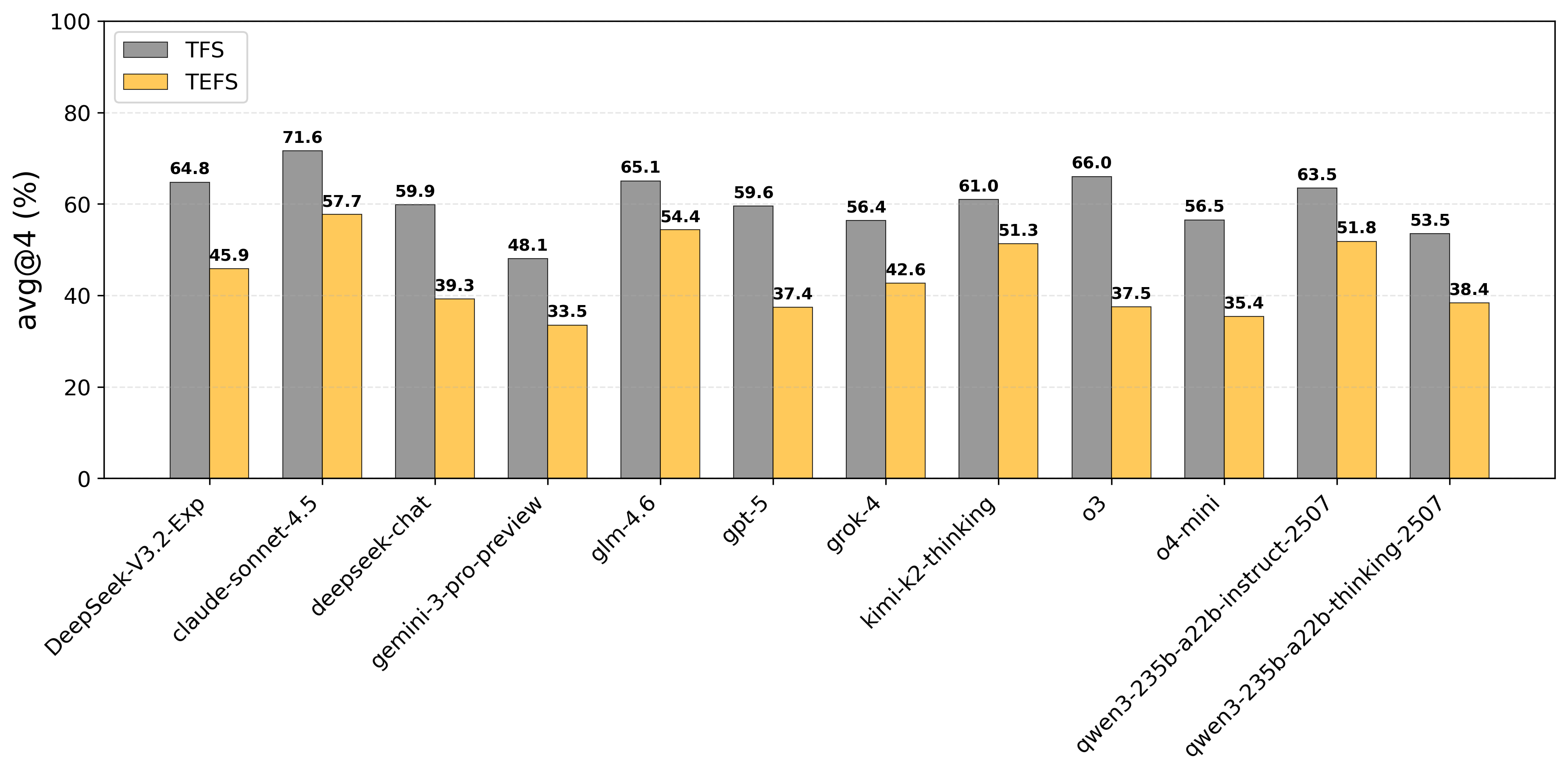}
    \caption{Evaluation Results of TFS and TEFS.}
    \label{fig:tfsTefs}
\end{figure*}

\bbb{Step 3}: Matching and Curation. 
This step is critical for constructing effective test cases. 
This step identifies potential "Task-Tool" pairs by matching similar tags. However, original task descriptions and tool definitions seldom align perfectly. Therefore, strict manual curation is performed, aiming to align the Task and Tool to achieve a "unique solution" with minimal modifications. 141 MCP tools are finally selected and are reconstructed into 141 MCP servers with one tool in each server.

\bbb{Step 4}: MCP Tool Code Generation. 
Finally, to enable automated evaluation in the sandbox, executable mock code for the MCP tools is required. 
LLM (e.g., GPT-4o) automatically generate Python stub functions based on the curated tool definitions (including tool name, description, and parameters). As automatically generated code may contain defects, an expert team reviews and modifies each function. This review ensures function signatures are identical to the definitions, the logic is correct, and the code executes safely within the sandbox.

This process yields a complete test case, which comprises three core components: (1) the task description, (2) the MCP tool definition and its verified mock code, and (3) the unique solution for the task.
These test cases are stored locally in JSON format, enabling dynamic loading within the sandbox environment during the evaluation process.

\subsection{Task Classification}
To assess the model's capabilities in various scenarios, we classified all tasks based on their complexity and domain specificity.

Task Domain: 
\bbb{1) Daily Tasks}: Covers daily-life scenarios such as entertainment and office work.
\bbb{2) Professional Tasks}: Involves specific domains, such as academic research or software engineering.

Invocation Complexity:
\bbb{1) Single-Tool Invocation}: The task can be resolved by invoking only one MCP tool. 
This tests the Agent's foundational ability to understand and select the correct tool.
\bbb{2) Dual-Tool Parallel Invocation}: The task requires the Agent to plan and invoke two independent tools concurrently. 
This assesses the Agent's task decomposition and parallel planning capabilities.
\bbb{3) Dual-Tool Serial Invocation}: The task requires the Agent to invoke tools in two sequential steps, following a specific logical order. 
For example, the tool invocation in the second step may depend on the output from the first. This assesses the Agent's capabilities for multi-step reasoning, planning, and state maintenance.
\bbb{4) Multi-Tool Invocation}: The task requires the Agent to invoke tools in multiple steps according to a logical sequence. 
This may involve a combination of parallel and serial invocations, representing a more complex tool-use scenario.

Every Domain contains 30 tasks of Single-tool Invocation, plus 20 tasks of each other complex type of invocation (except the professional multi-tool invocation, which contains 18 tasks), for a total of 178 tasks.

\section{Evaluation}

\begin{table*}[h]
    \centering
    \caption{TFS avg@4 by Task Category}
    \label{tab:tfs_category}
    
    \resizebox{\textwidth}{!}{
\begin{tabular}{lcccccccc}
        \toprule
        \multirow{2}{*}{\textbf{Models}} & \multicolumn{4}{c}{\textbf{Daily}} & \multicolumn{4}{c}{\textbf{Professional}} \\
        \cmidrule(lr){2-5}\cmidrule(lr){6-9}
         & Single & Dual Serial & Dual Parallel & Multi & Single & Dual Serial & Dual Parallel & Multi \\
        \midrule
        \textbf{claude-sonnet-4.5} & 96.67 & 86.25 & 93.75 & 67.50 & 90.00 & 58.75 & 68.75 & 40.28 \\
        \textbf{DeepSeek-V3.2} & 91.67 & 58.75 & 68.75 & 56.25 & 89.17 & 50.00 & 62.50 & 36.11 \\
        \textbf{gemini-3-pro-preview} & 72.50 & 50.00 & 52.50 & 38.75 & 67.50 & 40.00 & 48.75 & 37.50 \\
        \textbf{glm-4.6} & 87.50 & 92.50 & 77.50 & 50.00 & 83.33 & 58.75 & 61.25 & 44.44 \\
        \textbf{gpt-5} & 90.83 & 77.50 & 73.75 & 38.75 & 82.50 & 48.75 & 56.25 & 41.67 \\
        \textbf{grok-4} & 93.33 & 55.00 & 67.50 & 40.00 & 83.33 & 43.75 & 62.50 & 37.50 \\
        \textbf{kimi-k2-thinking} & 90.83 & 70.00 & 76.25 & 57.50 & 86.67 & 47.50 & 56.25 & 34.72 \\
        \textbf{o3} & 95.83 & 80.00 & 85.00 & 56.25 & 87.50 & 48.75 & 71.25 & 36.11 \\
        \textbf{o4-mini} & 93.33 & 72.50 & 70.00 & 37.50 & 91.67 & 57.50 & 65.00 & 13.89 \\
        \textbf{qwen3-235b-a22b-instruct-2507} & 89.17 & 76.25 & 77.50 & 60.00 & 89.17 & 47.50 & 63.75 & 31.94 \\
        \textbf{qwen3-235b-a22b-thinking-2507} & 94.17 & 40.00 & 80.00 & 32.50 & 86.67 & 33.75 & 70.00 & 29.17 \\
        \midrule
        \textbf{Average} & 91.04 & 70.73 & 75.10 & 49.90 & 85.90 & 47.71 & 61.98 & 34.95 \\
        \bottomrule
    \end{tabular}
    } 
    \label{tlb:tfs}
\end{table*}

\begin{table*}[h]
    \centering
    \caption{TEFS avg@4 by Task Category}
    \label{tab:tefs_category}

    \resizebox{\textwidth}{!}{
        \begin{tabular}{lcccccccc} 
\toprule
        \multirow{2}{*}{\textbf{Models}} & \multicolumn{4}{c}{\textbf{Daily}} & \multicolumn{4}{c}{\textbf{Professional}} \\
        \cmidrule(lr){2-5}\cmidrule(lr){6-9}
         & Single & Dual Serial & Dual Parallel & Multi & Single & Dual Serial & Dual Parallel & Multi \\
        \midrule
        \textbf{claude-sonnet-4.5} & 96.67 & 51.25 & 93.75 & 55.00 & 90.00 & 33.75 & 68.75 & 15.28 \\
        \textbf{DeepSeek-V3.2} & 91.67 & 58.75 & 22.50 & 12.50 & 89.17 & 48.75 & 13.75 & 26.39 \\
        \textbf{gemini-3-pro-preview} & 72.50 & 28.75 & 48.75 & 20.00 & 67.50 & 17.50 & 47.50 &  5.56 \\
        \textbf{glm-4.6} & 87.50 & 75.00 & 76.25 & 35.00 & 83.33 & 50.00 & 48.75 & 23.61 \\
        \textbf{gpt-5} & 90.83 & 77.50 &  0.00 & 10.00 & 82.50 & 48.75 &  0.00 & 30.56 \\
        \textbf{grok-4} & 93.33 & 31.25 & 67.50 & 27.50 & 83.33 & 20.00 & 62.50 &  5.56 \\
        \textbf{kimi-k2-thinking} & 90.83 & 57.50 & 72.50 & 41.25 & 86.67 & 37.50 & 43.75 & 22.22 \\
        \textbf{o3} & 95.83 & 80.00 &  0.00 &  8.75 & 87.50 & 48.75 &  0.00 & 25.00 \\
        \textbf{o4-mini} & 93.33 & 72.50 &  0.00 & 10.00 & 91.67 & 57.50 &  0.00 & 11.11 \\
        \textbf{qwen3-235b-a22b-instruct-2507} & 89.17 & 66.25 & 73.75 & 25.00 & 89.17 & 37.50 & 60.00 & 20.83 \\
        \textbf{qwen3-235b-a22b-thinking-2507} & 94.17 &  8.75 & 80.00 & 13.75 & 86.67 &  3.75 & 70.00 &  5.56 \\
        \midrule
        \textbf{Average} & 91.04 & 57.81 & 47.08 & 24.06 & 85.90 & 36.77 & 36.46 & 17.59 \\
        \bottomrule
        \end{tabular}
    }
    \label{tlb:tefs}
\end{table*}
\subsection{Evaluation Metrics}

To evaluate the Agent's capabilities from multiple dimensions, we define the following evaluation metrics, which cover task completion, execution efficiency, and resource consumption.

Let $N$ be the total number of tasks in the test set, and $T_i$ be the $i$-th task.
Let $G_i$ be the solution for task $T_i$, defined as a sequence of $n_i$ standard tool invocations.
The weight of task $T_i$ is its number of standard invocations, $|G_i| = n_i$.
Let $P_i$ be the tool invocation sequence actually generated by the Agent for task $T_i$.

\bbb{Task Finish Score (TFS)}:
A task $T_i$ is considered "Finished" ($\text{IsFinished}(T_i) = 1$) if and only if the set of tool invocations generated by the Agent, $P_i$, is identical to the set of invocations in the golden solution $G_i$. This requires an exact match of all "tool names" and "parameters" (where applicable), but does not consider the invocation order. TFS is the weighted average score across all tasks.
\[
TFS = \frac{\sum_{i=1}^{N} \text{IsFinished}(T_i) \cdot |G_i|}{\sum_{i=1}^{N} |G_i|}
\]

\bbb{Task Efficiency Finish Score (TEFS)}:
A task $T_i$ is considered "Efficiently Finished" ($\text{IsEfficientlyFinished}(T_i) = 1$) if and only if two conditions are met: (1) The task is "Finished" ($\text{IsFinished}(T_i) = 1$), and (2) The Agent's generated tool invocation sequence $P_i$ exactly matches the golden solution $G_i$ in its serial and parallel execution order. TEFS is the weighted average score across all tasks.
\[
TEFS = \frac{\sum_{i=1}^{N} \text{IsEfficientlyFinished}(T_i) \cdot |G_i|}{\sum_{i=1}^{N} |G_i|}
\]

\bbb{Resource Efficiency}: 
We also record the Agent's resource overhead during task execution to evaluate its cost-effectiveness. The "Total Score" in these metrics refers to the total weighted score (e.g., the numerator in the TFS formula: $\sum \text{IsFinished} \cdot |G_i|$).

\begin{itemize}[leftmargin=*, labelsep=0.5em]
\item Token Efficiency: 
Measures the score obtained per 1k output tokens consumed.
\[
\text{Token Efficiency} = \frac{\text{Total Score}}{\sum_{i=1}^{N} \text{Output Tokens}_i}
\]

\item Time Efficiency: 
Measures the score obtained per minute of execution time.
\[
\text{Time Efficiency} = \frac{\text{Total Score}}{\sum_{i=1}^{N} \text{Time}_i}
\]
\end{itemize}

In subsequent tests, the "Total Score" used for efficiency calculations is the total weighted score derived from TEFS.


 \begin{figure*}[h]
    \setlength{\abovecaptionskip}{0.1cm}
    \setlength{\belowcaptionskip}{-0.35cm}
    \centering  
\includegraphics[width=0.85\linewidth]{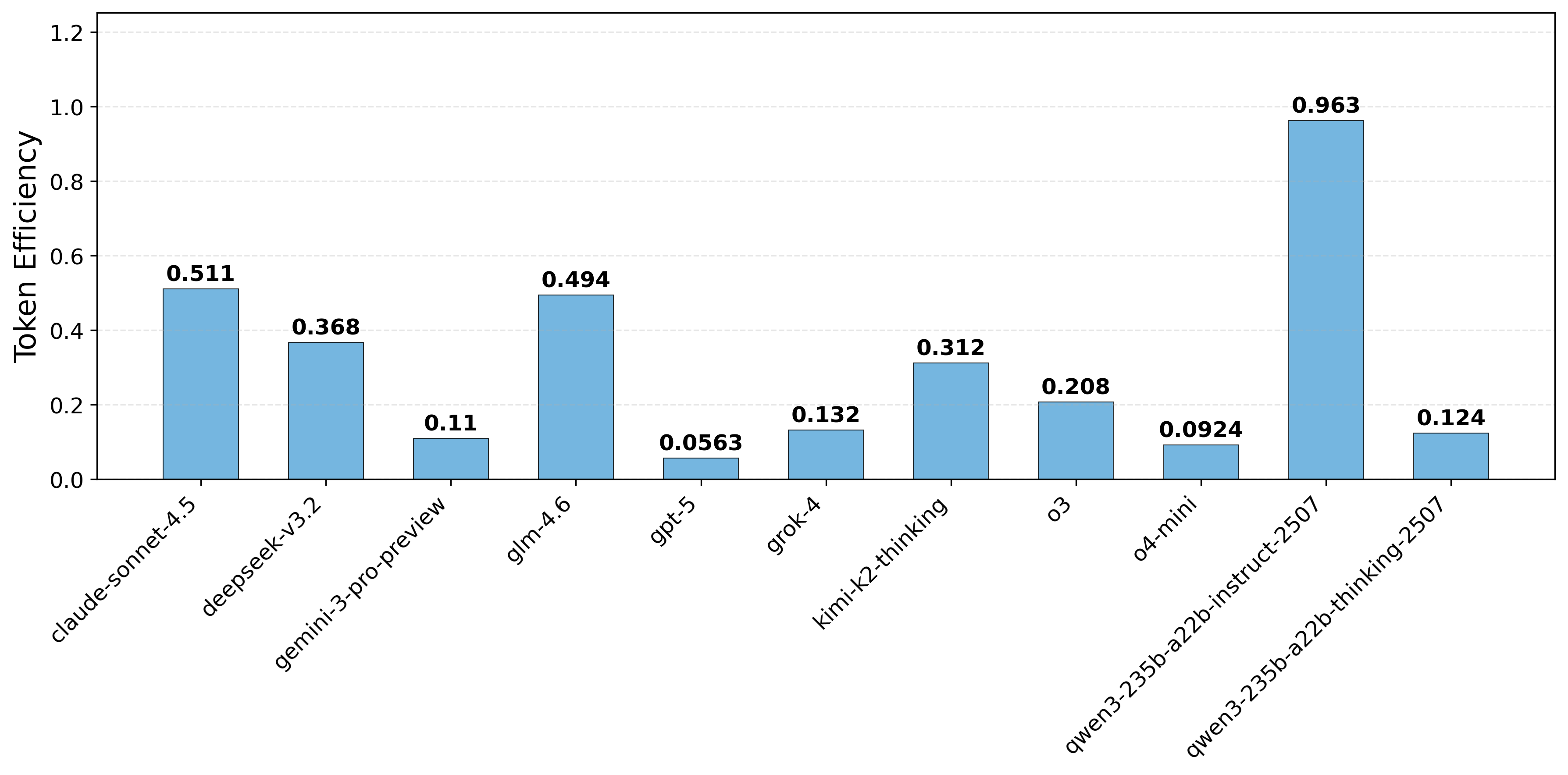}
    \caption{Evaluation Results of Token Efficiency.}
    \label{fig:tokenE}
\end{figure*}

 \begin{figure*}[h]
    \setlength{\abovecaptionskip}{0.1cm}
    \setlength{\belowcaptionskip}{-0.35cm}
    \centering  
\includegraphics[width=0.85\linewidth]{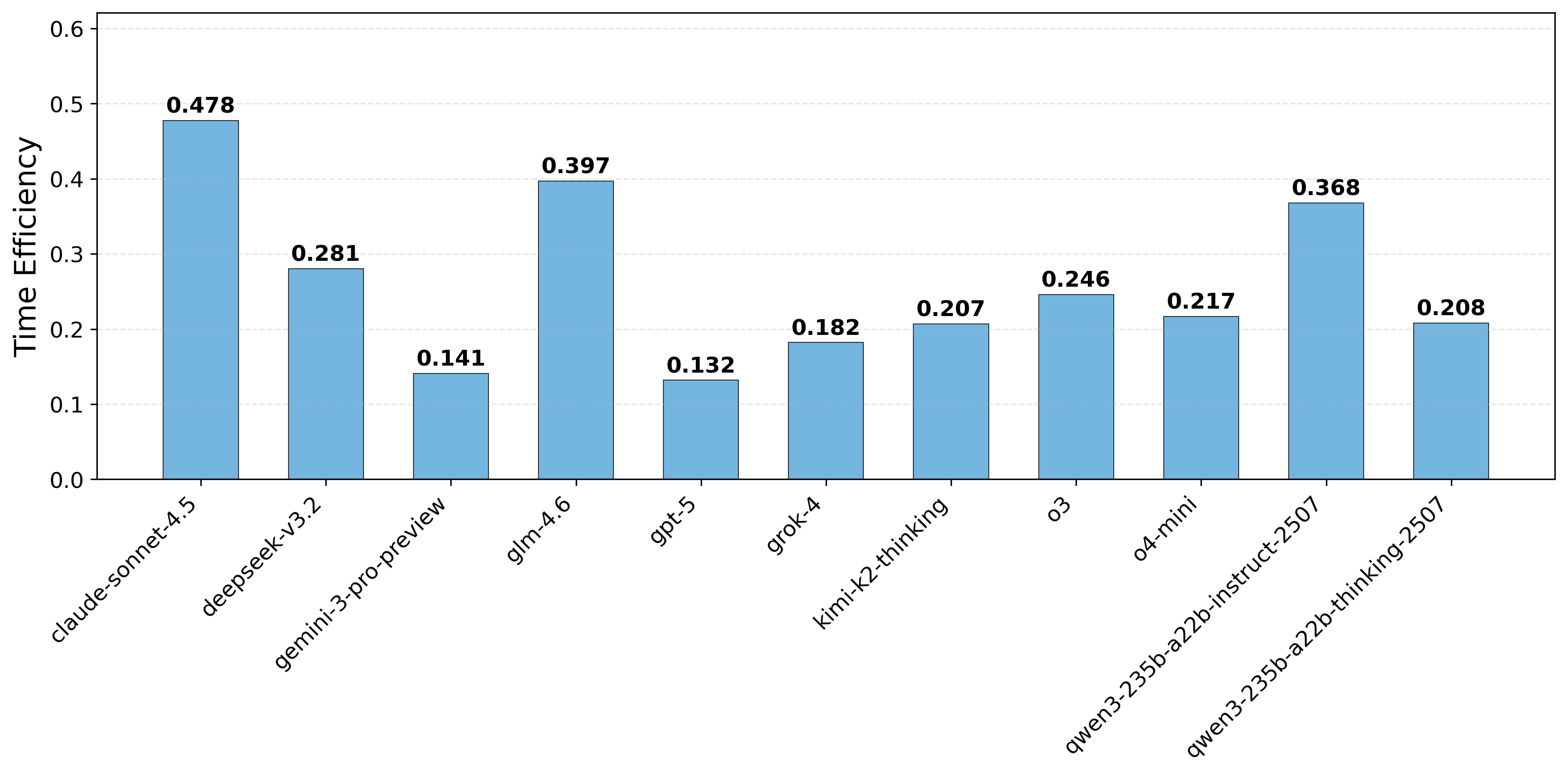}
    \caption{Evaluation Results of Time Efficiency.}
    \label{fig:timeE}
\end{figure*}

\subsection{Main Results}
We evaluate the performance of the following 11 mainstream models using the \system{} benchmark: Claude Sonnet 4.5 \cite{anthropic2025claudesonnet45}, DeepSeek V3.2 \cite{deepseekv3.2_2025}, Gemini 3 Pro Preview \cite{googleai2025gemini3propreview}, gpt-5 \cite{openai2025gpt5}, gpt-o3 \cite{openai2025o3}, gpt-o4-mini \cite{openai2025o4mini}, grok-4 \cite{xai2025grok4}, qwen3-235b-a22b-instruct-2507 \cite{yang2025qwen3}, qwen3-235b-a22b-thinking-2507 \cite{yang2025qwen3}, kimi-k2 \cite{bai2025kimiK2}, and glm-4.6 \cite{GLM46_2025}. For the performance comparison experiments, the number of tool invocation lists, $N$, is set to 20.

Figure \ref{fig:tfsTefs} presents the overall average TFS and TEFS scores (avg@4) for 11 mainstream models evaluated on \system{}. Under the TFS metric, Claude Sonnet 4.5, o3, and glm-4.6 achieve the top three scores of 71.6, 66.0, and 65.1, respectively, demonstrating superior task completion. In contrast, Gemini 3 Pro Preview records the lowest score of 48.1. Assessment under the stricter TEFS metric reveals that Claude Sonnet 4.5, glm-4.6, and qwen3-235b-a22b-instruct-2507 secure the top three positions in execution efficiency with respective scores of 57.7, 54.4, and 51.8, while Gemini 3 Pro Preview exhibits the lowest efficiency score of 33.5.

Nearly all evaluated models exhibit a decline of over 10 points in TEFS compared to TFS, with the o3 model recording the most significant drop of 28.5 points. This disparity indicates that current MCP tool-use capabilities prioritize task resolution over execution efficiency. Such neglect of efficiency results in substantial resource waste, specifically in terms of token consumption and execution time.

\begin{figure*}[h]
    \setlength{\abovecaptionskip}{-0.1cm}
    \setlength{\belowcaptionskip}{-0.15cm}
    \centering  

    \begin{minipage}[b]{0.45\linewidth}
        \centering
        \includegraphics[width=\linewidth]{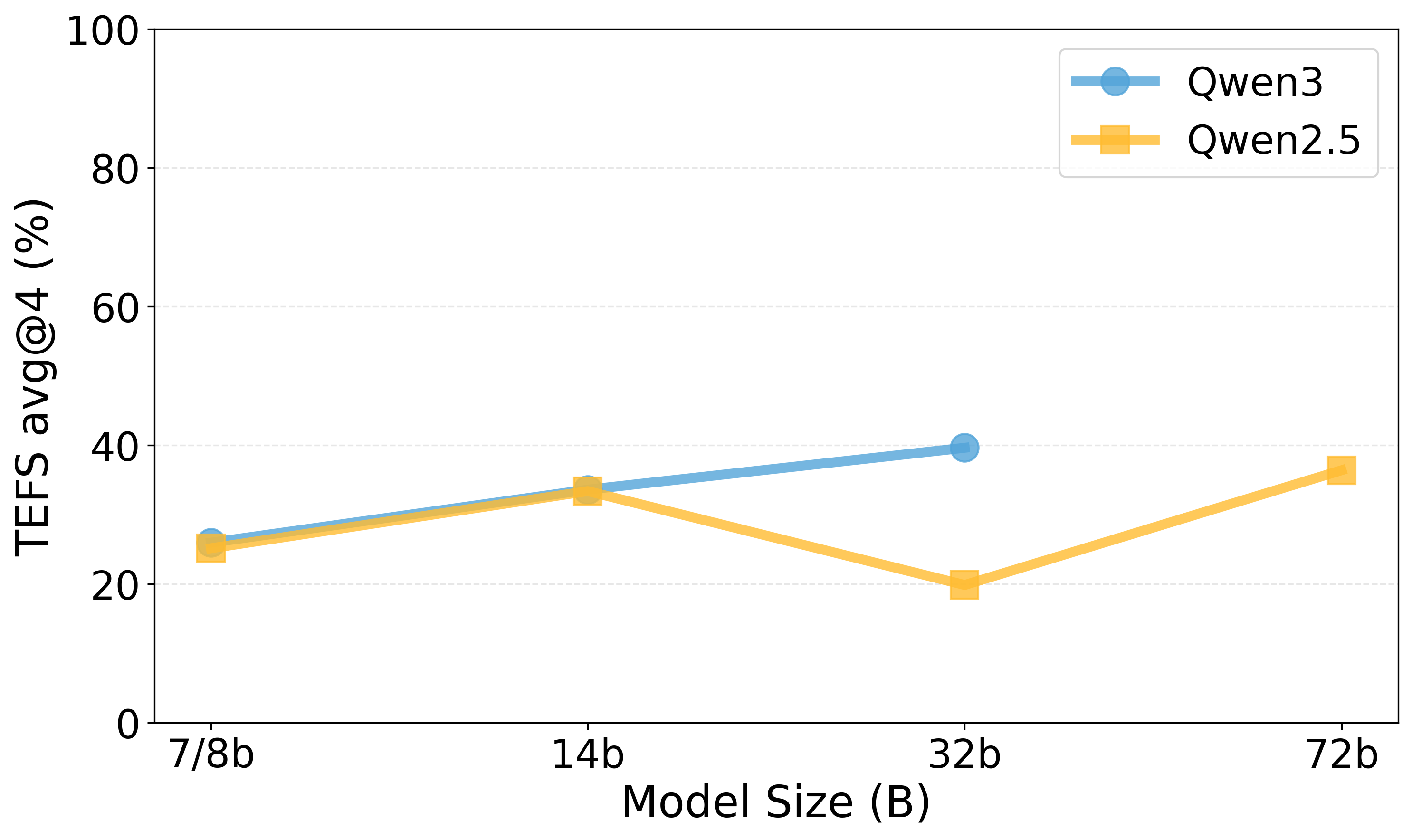}
        \small (a) Model Size v.s. TEFS Score
        \label{modelsize}
    \end{minipage}
    \begin{minipage}[b]{0.45\linewidth}
        \centering
        \includegraphics[width=\linewidth]{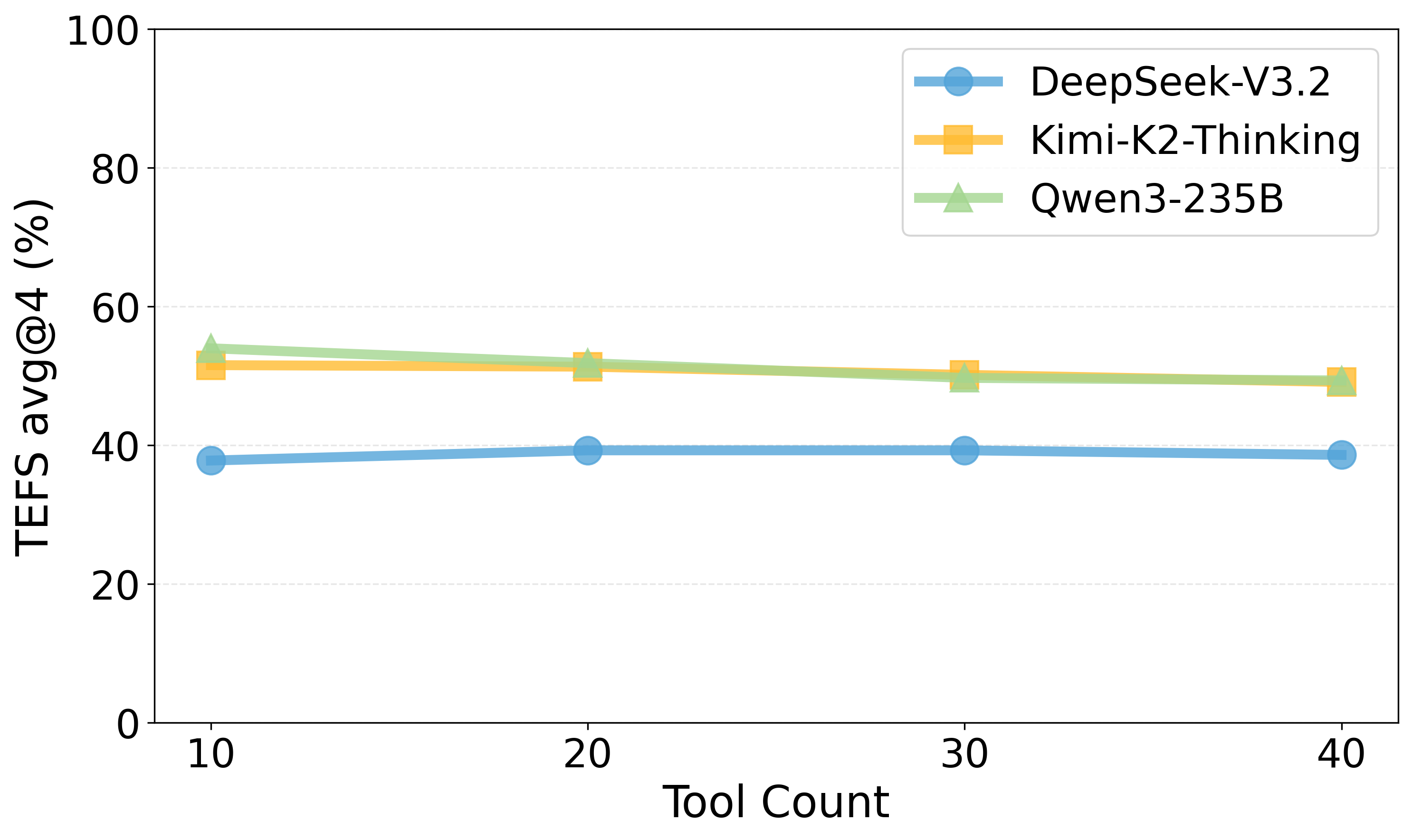}
        \small (b) Tool Count v.s. TEFS Score
        \label{toolcount}
    \end{minipage}
    \vspace{5pt} 
    \caption{The influence of model size and Tool Count on TEFS score}
    \label{TEFSInfluence}
\end{figure*}

Tables \ref{tlb:tfs} and \ref{tlb:tefs} present the scores across different task categories. Since TFS does not account for execution efficiency, it provides a more direct reflection of inherent task difficulty. In both Daily and Professional domains, average model scores decline as the number of tool invocations increases, showing that task difficulty scales from Single to Multi-tool scenarios. From a domain perspective, Professional tasks consistently exhibit higher difficulty than Daily tasks.

Furthermore, the average TFS for Dual Parallel tasks exceeds that of Dual Serial tasks, suggesting that, from a logic and completeness standpoint, serial tasks are more challenging to resolve. However, transitioning to the TEFS metric reveals a sharp and significant decline in Dual Parallel scores across all models. This trend indicates a widespread deficiency in correctly executing parallel tool calls. This limitation is particularly prominent in OpenAI series models (e.g., gpt-5), which record a TEFS of 0 for Dual Parallel tasks, failing to execute required parallel operations efficiently or accurately.

Further observation of task-specific performance reveals distinct differences in tool-invocation strategies among models. OpenAI models adopt an extreme serial approach, resulting in zero scores for Dual Parallel tasks. Conversely, Claude Sonnet 4.5 prioritizes parallelization wherever possible; while this leads to equivalent TFS and TEFS scores for Dual Parallel tasks, the model incorrectly applies parallel strategies to Dual Serial tasks, causing an anomalous drop in its TEFS score. Other models occupy an intermediate state between these two strategic extremes.

Figure \ref{fig:tokenE} illustrates the results for Token Efficiency. qwen3-235b-a22b-instruct-2507 exhibits the highest Token Efficiency, significantly higher than Claude Sonnet 4.5 and glm-4.6, which rank second and third, respectively. This leading performance is attributed to qwen3-235b-a22b-instruct-2507's highest score under the TEFS metric combined with its "NoThinking" design. Conversely, gpt-5 records the lowest Token Efficiency, suggesting that the excessive "thinking" tokens generated by gpt-5 do not translate into effective scores.

Figure \ref{fig:timeE} illustrates the results for Time Efficiency. Claude Sonnet 4.5 achieves the highest Time Efficiency, which is consistent with the earlier analysis of its aggressive parallel strategy selection. glm-4.6 and qwen3-235b-a22b-instruct-2507 rank second and third, respectively, while gpt-5 records the lowest Time Efficiency. These results are subject to external factors such as network latency and regional variations, which may lead to discrepancies during reproduction. To ensure result stability, the evaluation utilizes official APIs wherever possible.

Overall, the evaluated models demonstrate varying capabilities across different benchmarking metrics. Claude Sonnet 4.5 achieves the top ranking in TFS, TEFS, and Time Efficiency, underscoring its superior proficiency in MCP tool use. Specifically, its aggressive parallel tool-invocation strategy yields distinct advantages in both execution speed and performance on parallel tasks. Conversely, other models—most notably the OpenAI series—exhibit a significant deficit in parallel tool-use capability, resulting in substantial score reductions under the stricter TEFS evaluation dimension.

\subsection{Performance Analysis}

We further investigate the influence of model size and the number of candidate tools on TEFS.

First, we examine the change in avg@4 TEFS for different size models within the Qwen2.5 and Qwen3 series, setting the number of candidate tools to 10. As shown in Figure \ref{TEFSInfluence}(a), TEFS generally exhibits an upward trend as the model size increases. However, a noticeable dip in performance occurs at the Qwen 2.5 32B model, which may relate to its specific training methodology.

Next, we evaluate the impact of varying the number of candidate tools using Deepseek-V3.2, Kimi-K2-Thinking and Qwen3-235B. Figure \ref{TEFSInfluence}(b) presents the results. Although Deepseek-V3.2 shows a slight upward trend when the number of tools is 10 and 20, overall, as the number of alternative tools increases, the TEFS of all models show a slight downward trend.

In summary, the TEFS score generally increases with model scale and decreases as the number of distractor tools grows.
\section{Conclusion}
In this paper, we propose \system{}, an Autogen-based evaluation framework designed to measure the efficiency of large language models' MCP tool invocation for task completion. The framework constructs daily and professional tasks covering single-tool, dual-tool (serial or parallel), and multi-tool invocations by matching hundreds of collected tasks with \premcptoolnum{} MCP tools. The design of novel task completion efficiency metrics achieves automated evaluation of model capabilities. The relevant code is open-source.


\clearpage
\bibliography{reference}

\end{document}